\documentclass[a4paper, 10pt, conference]{ieeeconf}      
\usepackage{FG2024}

\FGfinalcopy 

\IEEEoverridecommandlockouts                              
\usepackage{subfigure}

\usepackage{amsmath,amsfonts}
\usepackage{algorithmic}
\usepackage{algorithm}
\usepackage{array}
\usepackage[caption=false,font=normalsize,labelfont=sf,textfont=sf]{subfig}
\usepackage{textcomp}
\usepackage{stfloats}
\usepackage{url}
\usepackage{verbatim}
\usepackage{graphicx}
\usepackage{cite}
\usepackage{bbm}
\usepackage{multirow}
\usepackage{textcomp}
\usepackage{pifont}
\usepackage{amssymb}
\usepackage{dsfont}

\def\FGPaperID{143} 

\title{\LARGE \bf
QGFace: Quality-Guided Joint Training for Mixed Quality Face Recognition
}

\author{\parbox{16cm}{\centering
    {\large Youzhe Song and Feng Wang}\\
    {\normalsize
    School of Computer Science and Technology, East China Normal University, Shanghai, China\\
    }}
}

\usepackage{fancyhdr}
\begin{document}

\ifFGfinal
\thispagestyle{empty}
\pagestyle{empty}
\else
\author{Anonymous FG2024 submission\\ Paper ID \FGPaperID \\}
\pagestyle{plain}
\fi
\maketitle
\thispagestyle{fancy}

\begin{abstract}

The quality of a face crop in an image is decided by many factors, such as camera resolution, distance, and illumination condition. This makes the discrimination of face images with different qualities a challenging problem in realistic applications.
However, most existing approaches are designed specifically for high-quality (HQ) or low-quality (LQ) images, and the performances would degrade for the mixed-quality images.
Besides, many methods ask for pre-trained feature extractors or other auxiliary structures to support the training and the evaluation.
In this paper, we point out that the key to better understand both the HQ and the LQ images simultaneously is to apply different learning methods according to their qualities.
We propose
a novel quality-guided joint training approach for mixed-quality face recognition, which could simultaneously learn images of different qualities with a single encoder.
Based on quality partition, the classification-based method is employed for HQ data learning. Meanwhile, for LQ images that lack identity information, we learn them with self-supervised image-image contrastive learning.
To effectively catch up the model update and improve the discriminability of contrastive learning in our joint training scenario, we further propose a proxy-updated real-time queue to compose the contrastive pairs with features from the genuine encoder.
Experiments on the low-quality datasets SCface and Tinyface, the mixed-quality dataset IJB-B, and five high-quality datasets demonstrate the effectiveness of our proposed approach in recognizing face images of different qualities.

\end{abstract}

\section{INTRODUCTION}
Existing face recognition (FR) methods have achieved near perfect performance on high-quality (HQ) face images~\cite{arcface, cosface, sphereface}. However, their performances significantly degrade when being applied to low-quality (LQ) images.
Some works focus on face recognition, where LQ images are adopted for matching between the given probe images and the gallery images~\cite{towards,id-aware,id-preserve,tcn,tc,distill}. However, due to the domain gap between the HQ and the LQ images, the discriminability for HQ images is somewhat reduced.
It is difficult to effectively learn from both LQ and HQ images simultaneously.
In reality, the quality of a face image used for recognition is influenced by numerous factors, such as camera resolution, distance, occlusion, and illumination condition. For instance, a high-resolution smart phone camera could produce low-quality face crops when there are too many faces in the picture or the photographer focuses on the scenery in the background.
Meanwhile, in surveillance situations, the image qualities also vary when the distance is changing (the setting in the SCface~\cite{scface} dataset).
Domain gaps also exist among these low-quality images.
In some limited situations where the image qualities are fixed, it is possible to select the appropriate model according to the image qualities. However, for applications such as photo albums and complicated surveillance systems with various cameras, the quality of face images vary considerably. Existing methods that focus only on HQ or LQ images are not suitable for such scenarios.

The problems with mixed-quality face recognition using a single feature extractor lie in the following aspects.
First, due to the domain gap between the HQ and the LQ data, it is hard to learn from and inference with images of different qualities simultaneously. Most existing methods emphasize either the HQ or the LQ images, while miss the other one.
Some quality-invariant methods~\cite{tcn, distill, tc, ran, ddl} attempt to understand the LQ images with the guidance of a pretrained HQ model.
This explicitly teaches the model to lean to the LQ images and thus the performance degrades for the HQ images. Besides, as the HQ images are encoded by a fixed model, the compromise between images of difference qualities is also under-optimized.
Second, we argue that there is a semantic problem when applying the classification-based method on LQ images.
The identities of faces are hard to predict due to the natural variations of their poses, ages, and makeups for the HQ images.
Compared with the HQ images, the LQ images typically have fewer details, which makes it challenging for the model to accurately classify them. This confusion arises because the model may struggle to differentiate between natural variations in the images and disturbances caused by the lower image qualities.
As a result, LQ images usually stumble the performance of the feature extractor trained with classification loss~\cite{adaface}.

To address the above problems, we propose a novel approach, namely quality-guided joint training for mixed-quality face recognition (QGFace).
Our goal is to learn both the HQ and the LQ face images simultaneously, and to recognize face images without distinction of qualities with a single feature extractor.
To effectively utilize the data of different qualities, we design a quality partitioning strategy and apply different supervision signals according to the qualities of the images.
Specifically, for the HQ images, the classification-based method is applied as they have achieved near-perfect performance for face recognition~\cite{cosface, adaface, arcface, yang_orthogonality_2021}.
For the LQ images that could mislead the classification loss, we adopt self-supervised contrastive learning, which learns the face images at the instance level and alleviates the requirements for the image details.
By taking advantage of the classification methods and contrastive learning, we get an encoder which is applicable for images with different qualities.
Furthermore, with the joint training strategy, we only need a single feature extractor during training and evaluation. This makes our method highly extensible, which could serve as the base component for future works.

Nevertheless, taking contrastive learning from the unsupervised learning into the joint training scenario is non-trivial.
Face identification asks for searching the related identity of a given image. Classification-based methods contain thousands of image-identity pairs to learn discriminative features which could meet the above requirement.
For contrastive learning, it usually takes a large number of sample pairs to improve stability and accuracy. Momentum queue is a common choice~\cite{moco}. It generates representations by an extra encoder which is momentum-updated with the training encoder. However, it loses track of the model which is also guided by the classification supervision signal, and the produced features are whitened as the encoder keeps an average of all the past model statuses.
Inspired by BroadFace~\cite{broadface}, we propose a novel proxy-updated real-time contrastive queue to supply positive pairs from the genuine model with a large number of negative samples which is dynamically updated. The proxy is represented by the corresponding weights of an identity in the classifier.

The contributions of this paper are summarized as follows:

\begin{itemize}
 \item We propose a quality-guided joint training strategy for face recognition (QGFace) on the mixed-quality images. To better understand the LQ images that could mislead the classification methods, we partition the images into the HQ and LQ subsets.
 The classification loss and contrastive loss are applied in the two subsets, respectively. In this way, we can learn comprehensively from both the LQ and the HQ faces to train a single encoder.
 \item To support feature search in face identification, we propose a proxy-updated real-time contrastive queue to ensure the discriminability of the LQ images learned by contrastive learning.
 The identity proxy representations are taken to update the queue features encoded in the previous training step.
 \item We conduct extensive experiments on high, low, and mixed quality datasets. Our approach achieves competitive results on 
 different categories of datasets compared with the SOTA methods, which demonstrates the effectiveness of our QGFace.
\end{itemize}

\begin{figure*}
    \centering
    \includegraphics[width=1.8\columnwidth]{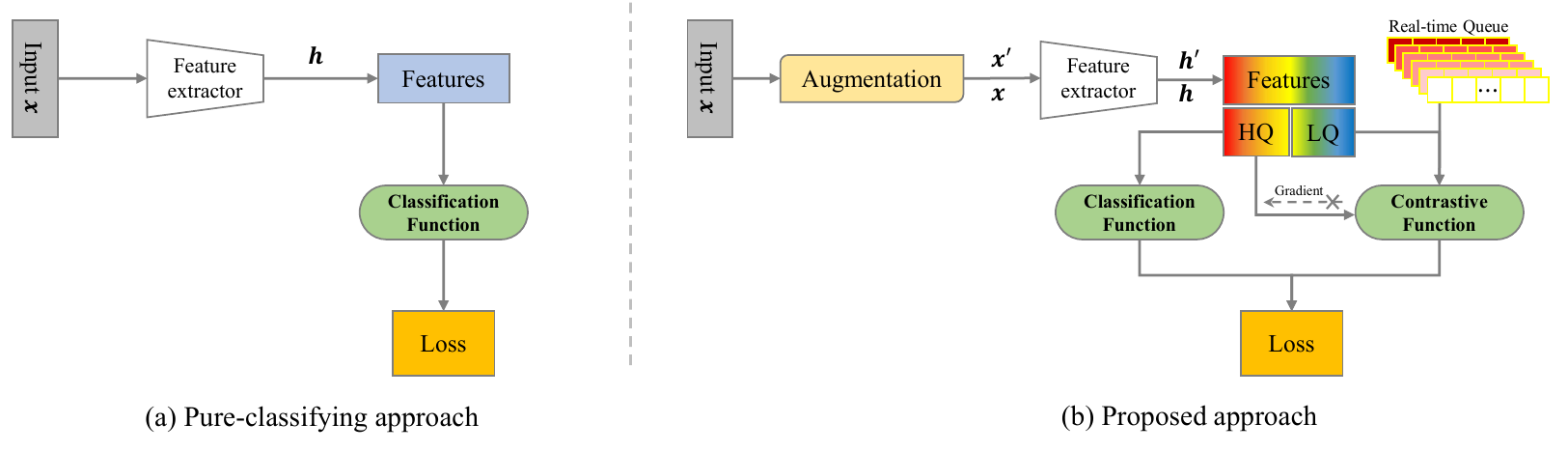}
    \caption{
    Conventional pure classification method vs. our QGFace.
    (a) An FR training pipeline with a single classification loss. All input images are taken into a single main loss.
    (b) Our pipeline additionally takes contrastive loss for low-quality images.
    We apply data augmentation to images and split their features into two parts: HQ and LQ ones. The HQ images are supervised by the classification loss, whereas the LQ images and their related HQ images are sent to the contrastive function. The gradient flow of the HQ data is stopped in contrastive learning. A real-time queue is designed to provide an effective feature queue and support the large-scale feature comparison.
    }
    \label{fig:framework}
\end{figure*}

\section{Related Works}

\subsection{Encoder-only methods}
Data augmentation is often used to produce LQ images, as they can provide domain information similar to the LQ evaluation datasets~\cite{distill,tc,tcn}.
However, these on-the-fly augmentation methods generally degrade the discriminability of images and make it difficult to identify them in model training.
Recent work discovers the relationship between the feature norm and image quality~\cite{magface, adaface}.
With awareness of the qualities of the image, the loss functions are designed to assign different importance to the images according to their qualities. They try to improve the performance of the LQ images while maintaining high accuracy on the HQ images.
However, the incompatibility between the LQ data and the classification training process prevents the encoder from effectively learning the hard LQ images.
PENB~\cite{penb} forces the alignment between the LQ and HQ data distribution from a global point of view, which may suffer from this problem.
Instead of learning LQ data with the classification-based method, we introduce contrastive learning to better understand them with the instance-level supervision. Furthermore, by quality partitioning, we can prevent LQ images from misleading classification-based HQ learning.

\subsection{Quality-invariant Methods}
To deal with the domain gap between the LQ and the HQ face images, some methods attempt to transfer the knowledge from the teacher model trained with the HQ images to the student model which focuses on the LQ images.
In~\cite{distill, tc}, a pretrained teacher network is frozen and then the student network is trained to extract features similar to the teacher with the down-sampled HQ images.
TCN~\cite{tcn} takes a contrastive loss to constrain the relationship between the HQ and the LQ images. 
RAN~\cite{ran} proposes multi-resolution Generative Adversarial Networks (GAN) to synthesize the cross-resolution images and a feature adaption network to enhance the communication between the feature extractors for the HQ and the LQ images.
DDL~\cite{ddl} divides the data into two subsets according to the predefined semantic difficulty, including yaw, race and distance during image capture. A similarity-based KL-divergence loss is applied to the two subsets.

These methods pursue the transfer of knowledge from the HQ to the LQ images. However, with such a single-way learning process, the performances of the learned models on the HQ images are stumbled.
Our QGFace partitions the training data according to their qualities and applies different supervising methods on them, which not only suits for the LQ face recognition, but also keeps the ability of recognizing the HQ images.

\subsection{Synthesis-based Methods}
To generalize the high performance of existing high-quality face recognition to different qualities, two straightforward ways are to super-resolve the LQ images for recognition~\cite{id-aware, id-preserve, super-id, towards} and to synthesize meaningful multi-quality training data. As they can only learn the mapping between the artificially downsampled images and the original one, the training of generation model usually needs a well-trained feature extractor.
In~\cite{id-aware, id-preserve, improve-super, super-id, towards, tinyface, tcsvt-lq, tcsvt-shr}, a pre-trained feature extractor is taken to supervise the identities of the synthetic images.
In~\cite{h2l}, a GAN is trained to learn how to downgrade high-resolution images to generate realistic low-resolution images. In~\cite{srdcr}, a DCR model is taken to preserve identity information, while in \cite{srgan, srda} the structure of the network is modified to improve the quality of the generation. Due to the absence of large-scale labeled LR face datasets, the approaches in \cite{cfsm, ideanet} synthesize low-quality images from the high-quality dataset and make it possible to train a mixed-quality FR model. \cite{tcsvt-lq} only focuses on a specific type of LQ images, which are partial face images. \cite{tcsvt-shr} reconstructs the HQ face images with singular value decomposition to improve reliability.

These methods could obtain satisfactory HQ images to mitigate the effect of resolution difference or lead the model to learn to discriminate LQ face images. However, since an additional generative module is needed, the overall framework is not end-to-end and computationally expensive.
Besides, a well-trained feature extractor is still needed in these methods, which highly dominates the training of the synthesis-based model and the performance of the recognition process. By directly training a single feature extractor which is applicable to both the HQ and the LQ data in an end-to-end way, our method is more elegant and orthogonal with the synthesis-based methods.

\section{Method}

Fig.~\ref{fig:framework}(b) illustrates the framework of our quality-guided joint training for mixed-quality face recognition.
We first partition the features into HQ and LQ subsets according to the image qualities, and learn them with classification loss and contrastive loss, respectively.
Then, we propose a proxy-updated real-time contrastive queue to support the face identification among LQ images.


\subsection{Quality Partitioning}
\label{sec:adaface}

Taking into account the domain gap between the HQ and the LQ images, we partition the mixed quality images into the LQ and HQ subsets during training. In~\cite{magface, adaface}, it is found that the feature norm is highly correlated with the image quality. In our approach, we use the feature norm as the quality indicator for data partitioning.

Let $z_i$ denote the norm of the feature $\boldsymbol{h}_i$,
and $\mu_z$, $\sigma_z$ denote the mean and the standard deviation of $z$ respectively. With these statistics, we could approach the normalized distribution of $z$.
\begin{align}
\widehat{z}_i &=\left\lfloor\frac{z_i-\mu_z}{\sigma_z/c}\right\rceil_{-1}^{1},
\end{align}
where $c$ is a scale parameter to keep most values of $\widehat{z}$ in (-1,1). The gradient flow of $\widehat{z}$ is stopped during backpropagation to keep the norm from being optimized directly.

We partition the mixed-quality input data into the HQ and LQ subsets based on the quality indicator $\widehat{z}$ using a quality threshold $b$.
Classification loss and contrastive learning are then applied for the HQ and the LQ feature learning, respectively.
We define dependent but different quality partitioning strategies for these two methods according to their inherent properties.
For classification loss $L_{class}$, each HQ feature is processed individually. We apply the quality comparison to the features in the same way, i.e., the features with the quality indicator $\widehat{z}_i$ higher than $b$ are fed into the classification loss.
The contrast loss $L_{contra}$ focuses on the relationship between the pair of samples. We take the representations of the augmented image and the original one as a positive contrastive pair ($q$, $k$) for learning. We define the quality indicator of the $p-{th}$ positive pair as the lower one of its two components. Pairs whose quality indicators are lower than the quality threshold $b$ are fed into contrastive learning. Specifically, 
\begin{align}
\widehat{z}_p & = \min(\widehat{z}_p^q, \widehat{z}_p^k),\\
\mathcal{L}_{classification} &= \left\{
             \begin{array}{lr}
             {L}_{class}, &\widehat{z}_i > b   \\
             0, &else
             \end{array}
\right. ,\\
\mathcal{L}_{contrastive} &= \left\{
             \begin{array}{lr}
             {L}_{contra}, &\widehat{z}_p \le b \\
             0, &else
             \end{array}
\right. ,\\
\mathcal{L} &= \mathcal{L}_{classification} + \mathcal{L}_{contrastive} .
\end{align}
The outcomes of the two kinds of losses after quality partitioning are denoted as $\mathcal{L}_{classification}$ and $\mathcal{L}_{contrastive}$ respectively. We take their summation $\mathcal{L}$ as the final loss.

\subsection{Quality-guided joint training}\label{sec:quality-guided}
The quality-invariant methods learn the LQ information from the corresponding features of the HQ images encoded by a pretrained model, and apply contrastive loss or the KL-divergence loss on all data. Differently from them, we guide the model to learn the HQ and the LQ images simultaneously by applying classification loss and contrastive learning on different features according to their qualities. In this way, the supervision compromise between the HQ and the LQ features is minimized, and we can make full use of images with different qualities to train the encoder for the mixed-quality face recognition task.
During the evaluation process, these two loss functions and their related structures will be eliminated and only a single feature extractor is reserved, which can process images without distinction of qualities to enable mixed-quality face recognition.

\subsubsection{Classification loss for HQ features}

Margin-based softmax loss functions are widely used in face recognition and have achieved SOTA performances on the HQ evaluation datasets~\cite{cosface, arcface, sphereface}. This kind of method adds a margin parameter to enlarge the distance between the matched image-identity pair and the unrelated ones to discriminate between different identities. Their loss could be shown as:
\begin{equation}
  {L} = - \log
  \frac{e^{s\cdot f(\boldsymbol{h}_i, \boldsymbol{W}_{y_i})}}
      {e^{s\cdot f(\boldsymbol{h}_i, \boldsymbol{W}_{y_i})} + \sum^{n}_{j=1, j \neq y_i}e^{s\cdot f'(\boldsymbol{h}_i, \boldsymbol{W}_j)}},
      \label{eq:softmax}\\
\end{equation}
where $f(\boldsymbol{h}_i, \boldsymbol{W}_{y_i})$ and $f'(\boldsymbol{h}_i, \boldsymbol{W}_j)$ are two different functions to modulate the positive and the negative pair production of the feature $\boldsymbol{h} \in \mathbb{R}^d$, $y_i$ denotes the corresponding identity of $\boldsymbol{h}_i$,
$\boldsymbol{W} \in \mathbb{R}^{d\times n}$ is the weight of the classifier with $d$ being the feature dimension and $n$ being the number of classes, and $s$ is a scale parameter.


AdaFace~\cite{adaface} introduces two norm-related margin parameters $g_{angle}$ and $g_{add}$ to leverage the quality indicator with a predefined static margin $m$, where
\begin{align}
g_{angle} &= -m \cdot \widehat{z}_i, \\
g_{add} &= m \cdot \widehat{z}_i + m, \\
f_{AdaFace}(\boldsymbol{h}_i, \boldsymbol{W}_{y_i}) &= \cos(\theta_{y_i} + g_{angle}) - g_{add}\label{eq:ada+}, \\
f'_{AdaFace}(\boldsymbol{h}_i, \boldsymbol{W}_{j}) &= \cos(\theta_j)\label{eq:ada-}.
\end{align}
With further analysis on the gradient equations of ${L}$, a gradient scaling term (GST) ${g}$ could be extracted to understand the effects of the cross-entropy softmax-based loss functions, 
\begin{align}
  P_{y_i}^{(i)} &=\frac{\exp\left(f\left(\cos\theta_{y_i}\right)\right)}{\exp\left(f\left(\cos\theta_{y_i}\right)\right)+\sum_{j=1,j\neq y_i}^{n}\exp\left(s\cos\theta_j\right)}, \\
 g &:=\left(P_{y_i}^{(i)}-1\right)\frac{\partial f(\cos\theta_{y_i})}{\partial\cos\theta_{y_i}},
\end{align}
where $P_{y_i}^{(i)}$ denotes the probability of an input $\boldsymbol{x}_i$ on its target class $y_i$. After substituting \ref{eq:ada+} and~\ref{eq:ada-} into \ref{eq:softmax}, we get the AdaFace loss, which is taken as our classification loss $L_{class}$ for the HQ feature learning.

\subsubsection{Contrastive learning for LQ features}

\begin{figure}
    \centering
    \includegraphics[scale=0.35]{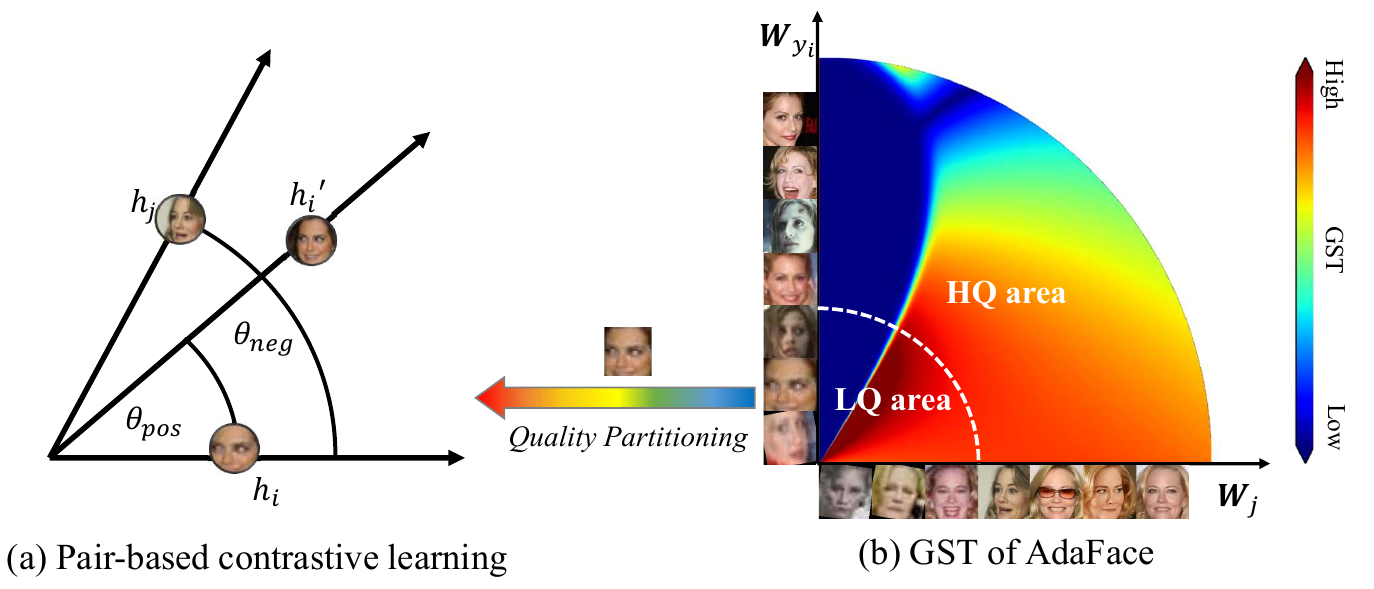}
    \caption{Illustration of contrastive learning for the LQ features.
    The distance between a data point and the origin denotes its feature norm.
    The GST shows that AdaFace puts a lot of attentions on the low-quality samples.
    The classification process matches the images with the abstract identity proxies, which is challenging when learning with the LQ data. We take the instance-level contrastive learning as shown in (a) on these images to relieve the learning burden.}
    \label{fig:gst}
\end{figure}

Although classification-based methods achieve SOTA performances in HQ face recognition, they are not suitable for learning LQ images. Fig.~\ref{fig:gst} visualizes the GST of AdaFace. 
It is obvious that LQ images gain more attention during training. However, due to the low qualities, they lack the discriminative information for matching with thousands of abstractive identity features.
To alleviate this problem, we apply contrastive learning on LQ features, which is based on the similarity between image instances and is more practical for learning the LQ data.

To comprehensively learn the LQ images and compose the positive contrastive pairs, we apply data augmentation on the HQ training images.
Unlike popular contrastive methods that apply data augmentation on images twice to compose the positive pairs, we only augment each image once to compose the pair with the original image. In addition, we stop the gradients of the HQ images during training to keep the qualities of their features. Let $(\boldsymbol{x}^q, \boldsymbol{x}^{k_+})$ denote the augmented image and the original image, respectively, and $(\boldsymbol{h}^q, \boldsymbol{h}^{k_+})$ denote their features, which compose a positive pair in contrastive learning.

Different augmentation methods are used, including scaling (downsampling to a lower resolution and then upsampling to the original size), crop \& random resize, rotation, and color distortion.
AdaFace introduces hard data augmentations with low possibilities including cropping 20\% of the original image, which makes the augmented images unrecognizable. This might be acceptable when little emphasis is put on these data, but is not suitable in our situation where all LQ features are used for learning. We increase the possibility of data augmentation with lower-level distortion. In addition, we follow \cite{towards} to add JPEG compression artifacts to make the augmented images more realistic.

Finally, to better cooperate with the above two learning paradigms, we employ the Supervised Contrastive Mask (SCM) introduced in CoReFace~\cite{coreface}. Pair-guided contrastive learning may pull two features away as they are not generated from the same image, while the classification method tries to pull them together as they are from the same identity.
To address this conflict, SCM removes all samples with the same identity of the given image from the negative feature pool. Let $\theta_{qk}$, $y_+$ denote the angle between feature pairs and the label of the given image, respectively, and $Q$, $k_j$ denote the number of negative samples and the $j^{th}$ feature among them. With a scale parameter $s$, the contrastive loss can be formulated as
\begin{equation}
f_{contra}(\boldsymbol{h}^q, \boldsymbol{h}^k) = \cos(\theta_{qk}).
\end{equation}
\begin{equation}
    {L}_{contra} = - \log 
\frac{e^{ s \cdot f_{contra}(\boldsymbol{h}^q, \boldsymbol{h}^{k_+})}}
    {\sum^{Q}_{j=1}\mathds{1}_{[y_j\neq y_+]}e^{ s \cdot f_{contra}(\boldsymbol{h}^q, \boldsymbol{h}^{k_j})}}.
\label{eq:contrastive}
\end{equation}


\begin{figure}
  \centering
  \subfigure[Proxy-updated Real-time Queue]{
  \includegraphics[width=0.6\columnwidth]{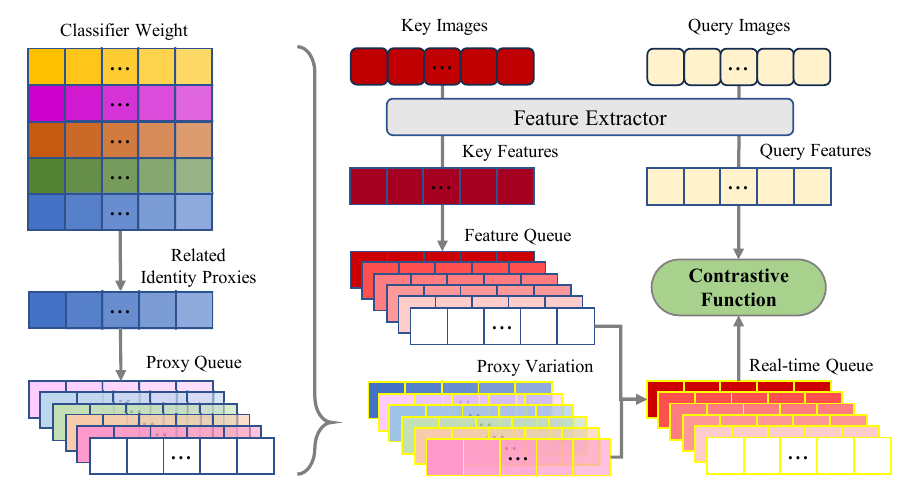}
  }
  \subfigure[Diff. between pairs]{
  \includegraphics[width=0.3\columnwidth]{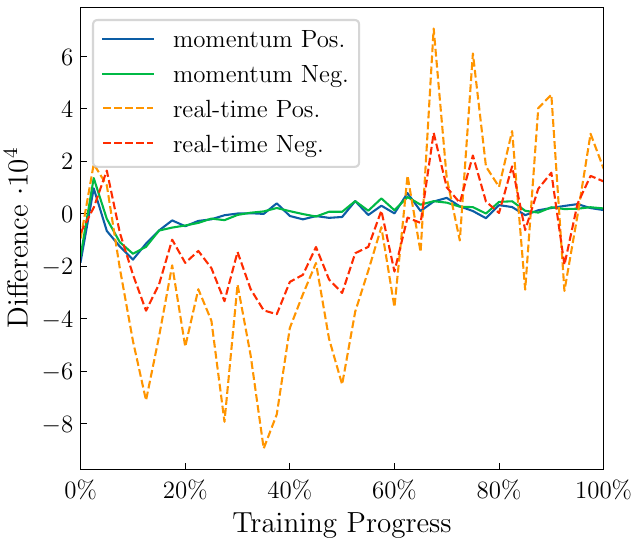}
  }
  \caption{(a) The updating process of our proposed proxy-updated real-time queue; (b) The difference between the positive and the negative pairs with different queues. The momentum queue faces a strongly limited boundary with the steady and whitened features. With our queue, contrastive learning can take the features from the training encoder to compose the positive pairs.
  }
  \label{fig:diff}
  \label{fig:queue}
\end{figure}

\subsection{Proxy-Updated Real-Time Queue}\label{sec:proxy-queue}

Self-supervised contrastive learning usually needs a large comparison pool to serve as negative candidates and push the training samples away from them.
MoCo~\cite{moco} defines the self-supervised contrastive learning as the comparison with a queue. A slowly progressing momentum encoder is proposed to produce keys and save them in a queue. The queries are produced by the training encoder. To make the positive and negative keys comparable, all of them are produced by the momentum encoder.
However, this operation prevents the training encoder from learning the LQ and HQ images simultaneously, as one of them is sent into a gradient-free encoder. Furthermore, as the momentum encoder is designed to be stable during large training intervals, this could lose track of the status of the encoder in joint training and produce whitened features.

To relieve the above problems, we propose a novel proxy-updated queue for contrastive learning, which is illustrated in Fig.~\ref{fig:queue}.
We take the genuine features encoded by the training model as positive pairs.
Since the training target of the cross-entropy softmax loss could constrain the corresponding feature-proxy pair to be similar, we update the negative features with their related proxies. Thus, we build a feature queue and a proxy queue.
Specifically, we first encode all the original images and the augmented ones. Then, we update the feature queue and the proxy queue with the above features and their corresponding identity proxies. As the features in queue could be generated from several iterations ago, we update them to be similar to the features generated by the training encoder. They are added with the difference between the proxies in queue and the proxies in training. To provide a mixed-quality queue, we take both $\boldsymbol{h}^q$ and $\boldsymbol{h}^k$ to update it. With SCM, no comparisons between identical features will be introduced. 

The cross-entropy softmax loss calculates the similarity between a feature and all identity proxies. To provide a comparative magnitude of pair relationship supervision, we set the size of the queue $Q$ as the number of the identities $n$. Thus, the number of pair comparisons used for contrastive learning keeps growing in training. This enables contrastive learning to discriminate among more images of mixed qualities for more effective learning.

\section{Experiments}

\begin{table*}
    \centering
    \caption{Ablations of our proposed framework. 
    ``HQ Avg.'' denotes the average performance on the five HQ datasets LFW, AgeDB, CFP-FP, CALFW and CPLFW. Rank-1 accuracy is reported for SCface and Tinyface. TAR@FAR=$10^{-4}$ is reported for IJB-B dataset.}
    \label{tab:abla}

    \begin{tabular}{c|c|c|c|c|c|c|c|c|c}
    \hline
    \multirow{2}{*}{\begin{tabular}[c]{@{}c@{}}Setting\end{tabular}}
    & \multirow{2}{*}{\begin{tabular}[c]{@{}c@{}}Augmentation\end{tabular}}
    & Quality
    & Contrastive
    & \multicolumn{3}{c|}{SCface}
    & \multirow{2}{*}{\begin{tabular}[c]{@{}c@{}}Tinyface\end{tabular}}
    & \multirow{2}{*}{\begin{tabular}[c]{@{}c@{}}IJB-B\end{tabular}}
    & \multirow{2}{*}{\begin{tabular}[c]{@{}c@{}}HQ Avg.\end{tabular}}
    \\
    \cline{5-7}
      &       &Partition    &Learning   & d1   & d2   &d3  & & &\\
    \hline
    \hline
    Baseline &\ding{55}  &\ding{55} &\ding{55} &61.69  &97.23  &99.85  &63.98 & 90.26 & 95.62 \\
    A &AdaFace  &\ding{55} &\ding{55}   &66.77  &99.08  &\textbf{100.0}  &67.49 & 74.07 &\textbf{95.76}\\
    B &AdaFace  &\ding{51} &\ding{55}   &68.15  &98.46  &\textbf{100.0}  &67.03 & 70.45 & 95.73 \\
    C &AdaFace  &\ding{51} &\ding{51}   &70.92  &98.62  &99.69  &66.58 &90.59 & 95.71 \\
    D &Ours  &\ding{55} &\ding{55}   &88.62  &99.23  &\textbf{100.0}  &68.80 &89.50 & 95.46 \\
    E &Ours  &\ding{51} &\ding{51}   &\textbf{92.31}&\textbf{99.54}  &\textbf{100.0}&\textbf{69.85}&\textbf{91.05}&95.43\\
    \hline
    \end{tabular}
\end{table*}

\subsection{Datasets and Implementation Details}

\textbf{Datasets.}
We take VGGFace2~\cite{vgg2} as the training dataset.
We evaluate our approach on three categories of datasets for different tasks. The first category is the HQ datasets including widely used LFW~\cite{huang_lfw_2007}, AgeDB~\cite{agedb}, CFP-FP~\cite{cfp-fp}, CPLFW~\cite{cplfw}, and CALFW~\cite{calfw}.
The second category is the LQ datasets including the SCface dataset~\cite{scface} and Tinyface dataset~\cite{tinyface} which are usually used for cross-resolution face recognition and LQ face recognition, respectively.
SCface contains 130 identities captured by five surveillance cameras at 3 distances: 4.2 m (d1), 2.6 m (d2) and 1.0 m (d3) for each identity. At each distance, the LQ images need to be matched with the HQ gallery images of the corresponding identities.
Tinyface contains 5K identities with 153.5K distracting images. The images in the gallery and the probe sets are of low resolution. This setting and the natural LQ property make it a challenging dataset.
For the third category, we choose IJB-B~\cite{ijbb} as the mixed-quality dataset which contains about 1.8K identities with a total of 21.8K images and 55K unconstrained video frames. In the 1:1 verification task, there are about 10K positive matches and 8M negative matches.

\textbf{Implementation Details}
We follow the settings commonly used in recent works~\cite{wang_mis-classified_2020,adaface,broadface,huang_curricularface_2020, magface} to crop and resize the face images to $112 \times 112$ with five landmarks~\cite{arcface}.
ResNet34~\cite{resnet} is employed as the backbone model. We take AdaFace as the classification loss. Our framework is implemented in Pytorch~\cite{pytorch}. We train the models on 4 NVIDIA A100 GPUs with a batch size of 512. All models are trained using SGD algorithm with an initial learning rate of $0.2$. We set the momentum to 0.9 and the weight decay to $5\times 10^{-4}$. The learning rate is divided by $10$ after the 6th and the 9th epochs, and the training stops after 12 epochs. The margin parameter $m$ is set to 0.4 as in AdaFace.
We set the scale parameter $s$ to 64 for both the classification loss and the contrastive loss.

\subsection{Ablation on different training strategies}


We investigate the effectiveness of the data augmentation strategy, the quality partitioning strategy (QP), and the contrastive learning (CL) with different categories of datasets in Table~\ref{tab:abla}.
We demonstrate the effectiveness of our method on the LQ and the HQ datasets, and its generalizability to the mixed-quality dataset by jointly learning the LQ and the HQ data simultaneously with the guidance of image quality.

\textbf{Baseline model:} We use the plain ResNet~\cite{resnet} model with AdaFace~\cite{adaface} loss function as our baseline. No augmentation is used except for the random horizontal flip.

\textbf{Setting A:} To show the influence of the LQ data, we apply the data augmentations recommended in AdaFace, i.e. down-sampling with a subsequent up-sampling, cropping with a small part, and random color distortion. All these three augmentations are applied with independent possibilities of 0.2.

\textbf{Setting B:} To show the negative influence of some extreme LQ data on the classification-based method, we apply quality partitioning on AdaFace loss simply by ignoring them (contrast learning is not presented).

\textbf{Setting C:} To show the effects of our proposed framework, we apply our QGFace with the AdaFace data augmentation. Both quality partitioning and contrastive learning are applied.

\textbf{Setting D:} To show the impacts of data augmentation, we apply our augmentation to the original AdaFace.

\textbf{Setting E:} Finally, we experiment our proposed method with quality partitioning, contrastive learning, and our data augmentation.

\begin{figure}
    \centering
    \includegraphics[width=0.8\columnwidth]{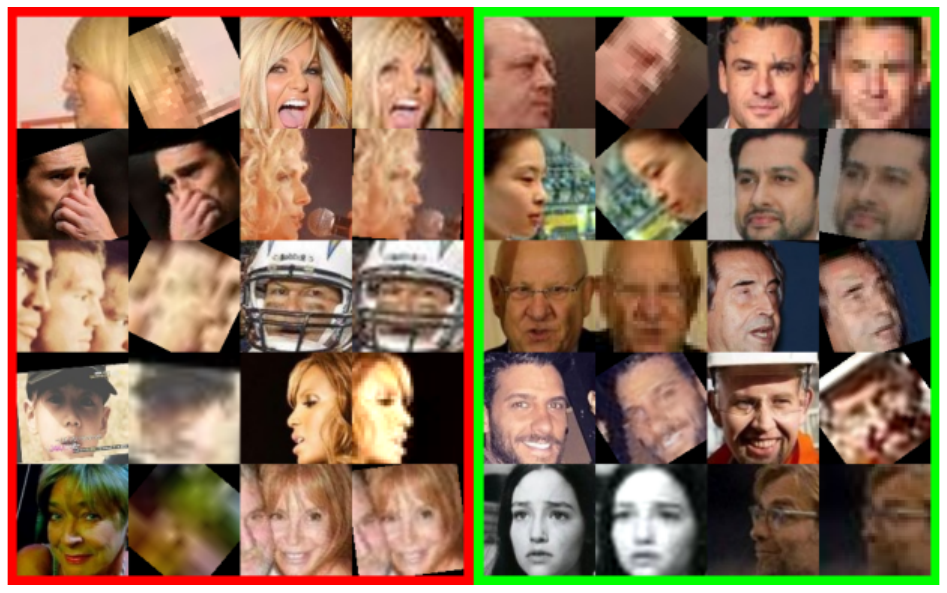}
    \caption{Examples of the image pairs. Every two columns show several pairs of original images and the augmented images. The pairs in red box contain LQ images while the pairs in green box are all HQ images. Our quality partitioning strategy is capable of distinguishing low-quality images which are blurred or with obstacles, or contain only a part of face.}
    \label{fig:aug}
\end{figure}

Table~\ref{tab:abla} shows the results of the above settings.
The baseline setting gains solid performance on the HQ datasets and IJB-B dataset. However, its performances on d1 of SCface dataset and Tinyface dataset are significantly poor.
In setting A, when data augmentation empowers the original feature extractor to gain higher performance on the HQ and the LQ datasets, the accuracy dramatically drops on the IJB-B dataset. This could be interpreted as a sacrifice of discriminability. With a single classification supervision signal, the features become less distinctive in a compromise between the HQ and the LQ images. On the IJB-B dataset where mixed quality matching between images requires stronger discriminability, this feature degradation misleads the comparisons among a large number of faces with various qualities.
Setting B shows that only dropping some LQ data helps the model gain better understanding on the LQ SCface dataset. This illustrates that some extremely low-quality data could mislead the classification method training. Even though AdaFace adapts the data importance according to the quality, it is harmful to learn these LQ data with classification loss.
In Setting C, our approach relieves the degradation caused by data augmentation in IJB-B and outperforms the baseline settings. Furthermore, the performance on SCface dataset is also improved. Meanwhile, our performance on the HQ data is only slightly dropped by less than 0.05\% compared with Setting B. This shows the effectiveness of contrastive learning on the LQ data and the ability of our method to keep the comprehension on the HQ data.
Setting D improves the performance on the LQ datasets and demonstrates the effectiveness of our data augmentation. Meanwhile, it is inferior to Setting C on IJB-B dataset which illustrates the effectiveness of our joint training framework again.

Finally, we implement Setting E, which is our proposed approach.
AdaFace learns hard-to-learn LQ images by classification loss with relatively less importance. Instead, we learn from them with contrastive learning. In this way, our model continues to learn the LQ images and gains more instance-level understanding from these less discriminative images. 
As a result, our method achieves the best results on almost all evaluation datasets of different qualities.
To ensure that most of the augmented LQ images are discriminable and realistic, we down-sample each image and then up-sample it to the original resolution. The discard ratio, the degree of brightness variations, and the contrast and saturation in color distortion are set to lower levels. In addition, we follow C. Kuo \textit{et al.}~\cite{towards} to add JPEG compression artifacts to make the image more realistic. With the new data augmentation, the augmented images become more understandable and compliant with the real-world low-resolution images.
Fig.~\ref{fig:aug} shows some examples. 
The image pairs in the red box are marked as LQ. They are harder to be identified than others in the training data (in green box) due to the blur, the obstacles, or the non-frontal positions in the images.
Our quality partitioning strategy sends these LQ data from the original and the augmented images to contrastive learning, and this setting outperforms all the other settings on SCface and Tinyface. Furthermore, it also shows promising results on the IJB-B dataset which contains a huge number of image pairs of mixed qualities.

\begin{figure}
    \centering
    \includegraphics[width=0.8\columnwidth]{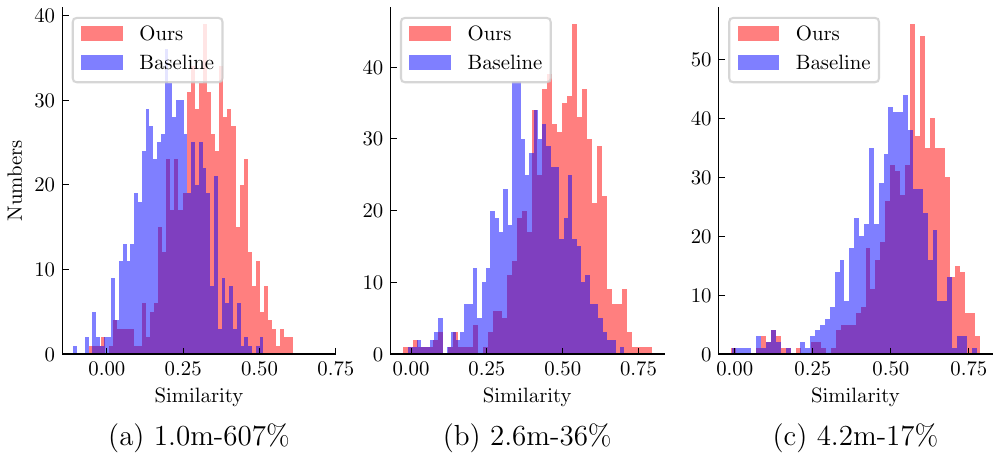}
    \caption{Similarity difference between QGFace and the Baseline on SCface. The histograms illustrate the similarity between probes and their related gallery images (matched pairs), where our method shows obvious advantages.
    The title of each sub-figure indicates the sub-dataset and the improvement of the difference on the mean similarity between the matched pairs and the most similar unmatched pairs.}
    \label{fig:similarity}
\end{figure}

\subsection{Ablation on the contrastive queue}

\begin{table}
\centering
\caption{Ablation of different queue settings.
``No'' indicates no contrastive loss. ``Batch'' indicates the samples in batch are taken as the negative pool.
}
\setlength{\tabcolsep}{1.0mm}

\begin{tabular}{c|c|c|c|c|c|c}
\hline
\multirow{2}{*}{\begin{tabular}[c]{@{}c@{}}Queue\end{tabular}}
& \multicolumn{3}{c|}{SCface}
& \multirow{2}{*}{\begin{tabular}[c]{@{}c@{}}Tinyface\end{tabular}}
& \multirow{2}{*}{\begin{tabular}[c]{@{}c@{}}IJB-B\end{tabular}}
& \multirow{2}{*}{\begin{tabular}[c]{@{}c@{}}HQ Avg.\end{tabular}} 
\\
\cline{2-4}
& d1 & d2 & d3 & & & \\
\hline
\hline
No           &87.85 &99.38 &\textbf{100.0}  &68.24 & 90.57 &\textbf{95.47}\\
Batch        &87.54 &99.38 &\textbf{100.0}  &68.67 & 87.38 & 95.28\\
Momentum     &63.08 &97.85 &99.85  &62.34 & 88.70 & 95.18\\
Proxy-updated  &\textbf{92.31} &\textbf{99.54} &\textbf{100.0}  &\textbf{69.85} &\textbf{91.05} & 95.43\\
\hline
\end{tabular}
\label{tab:queue}
\end{table}

Fig.~\ref{fig:diff} shows the average difference of the feature pairs with the momentum queue and our proxy-updated real-time queue. As the momentum queue is constructed with the features of a slowly progressing momentum encoder, the features are steady and whitened. Both the positive and the negative pairs are limited in a tight boundary. While in real scenarios where the difference is computed with the features from the same training encoder, it shows a much larger gap between the query and the corresponding key. Our proxy-updated queue takes these features as the positive pairs in contrastive learning directly and updates the features in queue with the identity proxies. According to the large size of the queue compared with the training batch, the difference between the negative pairs shows smaller magnitudes compared with the positive features in our queue.

Table~\ref{tab:queue} shows the performance of the models with different queue settings. In the ``No'' situation, contrastive learning is not applied, while the original and the augmented images are used for learning. In the ``Batch'' setting, the queue size is limited to the current batch. The number of feature-feature pairs in contrastive learning is extremely smaller than the feature-identity pairs in the classification loss. Thus, it stumbles on the IJB-B dataset which contains a large number of mixed-quality image pairs and needs higher discriminability.
The momentum queue utilizes whitened features to compose positive and negative pairs in contrastive learning. This misleads the optimization process in our joint training scenario. As a result, the performances of all datasets are lower than the setting without a queue.
Our proxy-updated queue provides genuine positive pairs and a dynamic negative queue to keep the same magnitude on the number of pair similarity computation as in the classification process. This enables contrastive loss to effectively learn from LQ images and maintain high discriminability. As a result, it outperforms the baseline and the other settings equipped with other queues on all benchmarks except a slight performance drop ($0.04\%$) on the HQ datasets.

\subsection{Ablation on the quality partitioning threshold}
\begin{table}
    \caption{Ablation of different quality partitioning thresholds. ``No'' indicates that the classification loss and the contrastive loss are simply added together to implement joint training.}
\centering
    \begin{tabular}{c|c|c|c|c|c|c}
    \hline
    \multirow{2}{*}{\begin{tabular}[c]{@{}c@{}}Threshold\end{tabular}}
    & \multicolumn{3}{c|}{SCface}
    & \multirow{2}{*}{\begin{tabular}[c]{@{}c@{}}Tinyface\end{tabular}}
    & \multirow{2}{*}{\begin{tabular}[c]{@{}c@{}}IJB-B\end{tabular}}
    & \multirow{2}{*}{\begin{tabular}[c]{@{}c@{}}HQ Avg.\end{tabular}}
    \\
    \cline{2-4}
    & d1 & d2 & d3 & & & \\
    \hline
    \hline
    No         &90.77 &99.69 &100.0  &69.80 &\textbf{91.26} &95.08 \\
    0.1        &86.00 &\textbf{99.54} &\textbf{100.0}  &68.13 & 90.87 & 95.41 \\
    0.2        &\textbf{92.31} &\textbf{99.54} &\textbf{100.0}  &\textbf{69.85} &91.05 &\textbf{95.43} \\
    0.3        &88.92 &\textbf{99.54} &\textbf{100.0}  &68.51 & 89.33 & 94.95 \\
    \hline
    \end{tabular}
    \label{tab:threshold}
\end{table}
Table~\ref{tab:threshold} shows the results of different quality partitioning thresholds. When no threshold is set, the classification loss and the contrastive loss supervise all the data at the same time. These two supervision signals guide the model to compromise between the features of the LQ and the HQ images, and the performances on SCface and HQ datasets are limited.
To alleviate the trivial parameter selection, we scale $\widehat{z}$ to $(0, 1)$ and thus the value of the threshold $b$ lies in the same range.
With a lower threshold, more LQ data is learned with the classification loss, and the performances on SCface and Tinyface drop. A larger threshold gives up the utilization of the labels of the relatively-high-quality images, which then results in degraded features. Finally, we choose $b=0.2$ in our implementation to balance the performances on the LQ and the HQ face images.

\subsection{Analysis on the SCface dataset}

Besides the quantitative analysis above, we further illustrate the difference between our QGFace and the Baseline on SCface in Fig.\ref{fig:similarity}. It is obvious that the matched pairs encoded by our QGFace are more similar than the Baseline method, which demonstrates the effectiveness of the quality partitioning strategy, the contrastive learning, and the selected data augmentation.
As the evaluation process aims to find an image among all of the candidates in the gallery, the most similar unmatched image is also important for the successful match. We then calculate the statistics of the similarity between a probe and its matched gallery image, and the similarity between the probe and its most similar unmatched gallery image. The difference between the means of the two kinds of similarity scores is calculated, and the improvement of our method is noted in the title of the sub-figures. When the distances become longer and the qualities of the images become lower, the improvement of QGFace becomes larger. In the case of 4.2m, the improvement is even more than 600\%, which clearly explains how our method improves the accuracy by more than 30\% on SCface-d1 in Table~\ref{tab:abla}.

\subsection{Comparison with SOTA methods}

\textbf{On SCface dataset.}
Table~\ref{tab:scface} compares our method with SOTA methods in SCface dataset.
Many quality-invariant methods need a fine-tuning process to learn the target domain information. This breaks the general end-to-end framework and limits the evaluation confidence as only 130 identities and no distracting faces are provided in SCface. Furthermore, all these methods show lower accuracy scores in d3 compared with the pure AdaFace trained with the HQ data. AdaFace-Aug, which is the same as Setting A in Table~\ref{tab:abla}, achieves 100.0\% by learning the relatively high quality augmented images.
This verifies our hypothesis that the quality-invariant methods lose discriminative information during training.
Our approach achieves the best performance on all three distances with clear margins compared with the other approaches. This demonstrates the effectiveness of our proposed quality-guided joint training strategy for mixed-quality face recognition by simultaneously learning both the HQ and the LQ images.

\textbf{On Tinyface dataset.}
Table~\ref{tab:tinyface} compares our method with other SOTA methods on Tinyface dataset. For the pure classification-based CurricularFace, even incorporated with a larger backbone model and a larger training dataset, it does not show an outstanding result compared with other methods. AdaFace puts relatively less emphasis on the hard LQ images and outperforms CurricularFace. However, it is still under-competitive with the methods that focus on LQ face recognition. Our method partitions the data according to the image qualities and learns the LQ images with contrastive learning instead of harder classification process. The approach in RIFR~\cite{distill} slightly outperforms our method by employing a distillation approach. This needs an extra pre-trained HQ face encoder to transfer its knowledge to the trained encoder, while our method does not have this preliminary requirement and is orthogonal with it. As we focus on training a feature extractor capable of dealing with mixed-quality images, it is easy to integrate QGFace with other methods to further improve performance. Furthermore, our method achieves a better average performance on SCface and Tinyface compared with RIFR~\cite{distill}, which demonstrates the effectiveness of our QGFace in recognizing mixed-quality face images under different scenarios.

\begin{table}
    \caption{Rank-1 accuracies (\%) of different methods on SCface.
    ``FT'' indicates that the method is fine-tuned. ``AdaFace-Aug'' is the same as setting A in Table~\ref{tab:abla}.
    }
    \centering
    \setlength{\tabcolsep}{1.0mm}

    \begin{tabular}{c|c|c|c|c}
    \hline
    Approach &Method   & d1 & d2 &d3  \\
    \hline
    \hline
    \multirow{10}{*}{\begin{tabular}[c]{@{}c@{}}Quality\\Invariant\end{tabular}}
    &T-C (IVC20)~\cite{tc}                          &45.10 &85.90 &96.10  \\
    &FAN (ACCV19)~\cite{fan}                         &62.00 &90.00 &94.00\\
    &RAN (ECCV20)~\cite{ran}                         &70.50 &96.00 &98.00 \\
    &S. -C. Lai \textit{et al.}(APSIPA ASC21)~\cite{sc}     &79.70 &95.70 &98.20 \\
    &FAN-FT(ACCV19)~\cite{fan}                      &77.50 &95.00 &98.30 \\
    &RAN-FT (ECCV20)~\cite{ran}                      &81.30 &97.80 &98.80 \\
    &DDL (ECCV20)~\cite{ddl}                         &86.80 &98.30 &98.30 \\
    &RIFR (TBIOM20)~\cite{distill}                  &88.30 &98.30 &98.60 \\
    &W. Hainan \textit{et al.} (FG23)~\cite{lrhr}     &89.00 &99.00 &99.25 \\
    &CATFace (TBIOM24)~\cite{catface}               &90.64 &98.85 &99.61\\
    \hline
    \multirow{3}{*}{\begin{tabular}[c]{@{}c@{}}Synthesis\end{tabular}}
    &SR-DCR (ICCPR20)~\cite{srdcr}   &74.10 &93.70 &97.2\\
    &M. Ullah \textit{et al.} (ICET21)~\cite{srgan}     &60.81 &90.45 &98.04\\
    &IDEA-Net (TIFS22)~\cite{ideanet}                 &90.76 &98.50 &99.25 \\
    \hline
    \multirow{5}{*}{\begin{tabular}[c]{@{}c@{}}Encoder-\\Only\end{tabular}}
    &NPT-Loss (TPAMI22)~\cite{npt}                    &85.69 &99.08 &99.08 \\
    &AdaFace (CVPR22)~\cite{adaface}                  &61.69 &97.23 &99.85 \\
    &AdaFace-Aug (CVPR22)~\cite{adaface}              &66.77 &99.08 &\textbf{100.0} \\
    &PENB-FT (AAAI23)~\cite{penb}                        &91.8  &99.0  &99.3 \\
    &\textbf{Ours}                  &\textbf{92.31} &\textbf{99.54} &\textbf{100.0} \\
    \hline
    \end{tabular}
    \label{tab:scface}
\end{table}

\section{Conclusion}
We have presented our approach of quality-guided joint training strategy for mixed-quality face recognition. 
Based on quality partitioning, the high-quality data is supervised with the classification loss while the low-quality data is supervised with the contrastive loss.
In this way, we can learn both the HQ and the LQ images simultaneously without any other network structures to cope with mixed-quality faces in real applications.
The proxy-updated real-time contrastive queue avoids the feature-whitening problem in the momentum encoder queue and improves the discriminability on the LQ data.
Our approach is as elegant as a single feature extractor, and it can be easily integrated with other approaches, such as super-resolution and distillation, to further improve the performance.

\begin{table}
    \caption{Rank-1 accuracy of different methods on Tinyface.
* means the method is trained with ResNet100 on MS1MV2 dataset.}
    \centering
    \begin{tabular}{c|c|c}
    \hline
    Approach &Method   & Rank-1  \\
    \hline
    \hline
    \multirow{9}{*}{\begin{tabular}[c]{@{}c@{}}Quality\\Invariant\end{tabular}}
    & QualNet50-LM (CVPR21)~\cite{qualnet}             &35.54\\
    & MobileFaceNet (ICPR21)~\cite{mobile}             &48.70\\
    & T-C (IVC20)~\cite{tc}                           &58.60\\
    & URL (CVPR20)~\cite{url}                          &63.89\\
    & RIFR (TBIOM20)~\cite{distill}                         &\textbf{70.40} \\
    & MIND-Resnet-50-FT (SPL21)~\cite{mindres}        &66.82\\
    & W. Hainan \textit{et al.} (FG23)~\cite{lrhr}     &66.3 \\
    & Z. Meng \textit{et al.}(TIP24)~\cite{texture}                        &54.3 \\
    & CATFacce* (TBIOM24)~\cite{catface}                        &68.95\\
    \hline
    \multirow{4}{*}{\begin{tabular}[c]{@{}c@{}}Synthesis\end{tabular}}
    & CSRI-FT (ACCV18)~\cite{tinyface}    &44.80\\
    & SRDA (ICoICT21)~\cite{srda}                  &34.15 \\
    & CFSM (ECCV22)~\cite{cfsm}            &63.01 \\
    & IDEA-Net~(TIFS22)~\cite{ideanet}     &68.13 \\
    \hline
    \multirow{5}{*}{\begin{tabular}[c]{@{}c@{}}Encoder-\\Only\end{tabular}}
    & CurricularFace* (CVPR20)~\cite{huang_curricularface_2020}    &63.68 \\
    & PENB-FT (AAAI23)~\cite{penb}         &39 \\
    & AdaFace~(CVPR22)~\cite{adaface}    &63.98 \\
    & AdaFace*~(CVPR22)~\cite{adaface}    &68.21 \\
    & \textbf{Ours}                       &69.85 \\
    \hline
    \end{tabular}
    \label{tab:tinyface}
\end{table}



{\small
\bibliographystyle{ieee}
\bibliography{egbib}

\begin{thebibliography}{10}\itemsep=-1pt

\bibitem{h2l}
A.~Bulat, J.~Yang, and G.~Tzimiropoulos.
\newblock To learn image super-resolution, use a gan to learn how to do image degradation first.
\newblock In {\em Proc. of ECCV}, 2018.

\bibitem{vgg2}
Q.~Cao, L.~Shen, W.~Xie, O.~M. Parkhi, and A.~Zisserman.
\newblock Vggface2: A dataset for recognising faces across pose and age.
\newblock {\em {IEEE} International Conference on Automatic Face \& Gesture Recognition, {FG}}, pages 67--74, 2018.

\bibitem{id-aware}
J.~Chen, J.~Chen, Z.~Wang, C.~Liang, and C.-W. Lin.
\newblock Identity-aware face super-resolution for low-resolution face recognition.
\newblock {\em IEEE Signal Processing Letters}, 27:645--649, 2020.

\bibitem{tinyface}
Z.~Cheng, X.~Zhu, and S.~Gong.
\newblock Low-resolution face recognition.
\newblock In {\em Asian Conference on Computer Vision}, 2018.

\bibitem{arcface}
J.~Deng, J.~Guo, N.~Xue, and S.~Zafeiriou.
\newblock Arcface: Additive angular margin loss for deep face recognition.
\newblock In {\em {IEEE} Conference on Computer Vision and Pattern Recognition, {CVPR}}, pages 4690--4699, 2019.

\bibitem{ran}
H.~Fang, W.~Deng, Y.~Zhong, and J.~Hu.
\newblock Generate to adapt: Resolution adaption network for surveillance face recognition.
\newblock In {\em Proc. of ECCV}, 2020.

\bibitem{scface}
M.~Grgic, K.~Delac, and S.~Grgic.
\newblock Scface – surveillance cameras face database.
\newblock {\em Multimedia Tools and Applications}, 51:863--879, 2011.

\bibitem{moco}
K.~He, H.~Fan, Y.~Wu, S.~Xie, and R.~B. Girshick.
\newblock Momentum contrast for unsupervised visual representation learning.
\newblock In {\em {IEEE} Conference on Computer Vision and Pattern Recognition, {CVPR}}, pages 9726--9735, 2020.

\bibitem{resnet}
K.~He, X.~Zhang, S.~Ren, and J.~Sun.
\newblock Deep residual learning for image recognition.
\newblock In {\em {IEEE} Conference on Computer Vision and Pattern Recognition, {CVPR}}, pages 770--778. {IEEE} Computer Society, 2016.

\bibitem{huang_lfw_2007}
G.~B. Huang, M.~A. Mattar, T.~L. Berg, and E.~Learned-Miller.
\newblock Labeled faces in the wild: A database for studying face recognition in unconstrained environments.
\newblock In {\em Tech. Rep.}, 2007.

\bibitem{ddl}
Y.~Huang, P.~Shen, Y.~Tai, S.~Li, X.~Liu, J.~Li, F.~Huang, and R.~Ji.
\newblock Improving face recognition from hard samples via distribution distillation loss.
\newblock In {\em Proc. of ECCV}, 2020.

\bibitem{huang_curricularface_2020}
Y.~Huang, Y.~Wang, Y.~Tai, X.~Liu, P.~Shen, S.~Li, J.~Li, and F.~Huang.
\newblock Curricularface: Adaptive curriculum learning loss for deep face recognition.
\newblock In {\em {IEEE} Conference on Computer Vision and Pattern Recognition, {CVPR}}, pages 5900--5909. {IEEE}, 2020.

\bibitem{tcsvt-shr}
M.~Jian and K.-M. Lam.
\newblock Simultaneous hallucination and recognition of low-resolution faces based on singular value decomposition.
\newblock {\em IEEE Transactions on Circuits and Systems for Video Technology}, 25(11):1761--1772, 2015.

\bibitem{distill}
S.~S. Khalid, M.~Awais, Z.~Feng, C.-H. Chan, A.~Farooq, A.~Akbari, and J.~Kittler.
\newblock Resolution invariant face recognition using a distillation approach.
\newblock {\em IEEE Transactions on Biometrics, Behavior, and Identity Science}, 2:410--420, 2020.

\bibitem{npt}
S.~S. Khalid, M.~Awais, Z.~Feng, C.-H. Chan, A.~Farooq, A.~Akbari, and J.~Kittler.
\newblock Npt-loss: Demystifying face recognition losses with nearest proxies triplet.
\newblock {\em IEEE Transactions on Pattern Analysis and Machine Intelligence}, 2022.

\bibitem{qualnet}
I.~Kim, S.~Han, J.~Baek, S.~Park, J.~Han, and J.~Shin.
\newblock Quality-agnostic image recognition via invertible decoder.
\newblock In {\em {IEEE} Conference on Computer Vision and Pattern Recognition, {CVPR}}, pages 12257--12266, 2021.

\bibitem{adaface}
M.~Kim, A.~K. Jain, and X.~Liu.
\newblock Adaface: Quality adaptive margin for face recognition.
\newblock In {\em {IEEE} Conference on Computer Vision and Pattern Recognition, {CVPR}}, pages 18729--18738. {IEEE}, 2022.

\bibitem{broadface}
Y.~Kim, W.~Park, and J.~Shin.
\newblock {BroadFace}: Looking at tens of thousands of people at once for face recognition.
\newblock In {\em Proc. of ECCV}, 2020.

\bibitem{towards}
C.~B. Kuo, Y.-T. Tsai, H.-H. Shuai, Y.-R. Yeh, and C.-C. Huang.
\newblock Towards understanding cross resolution feature matching for surveillance face recognition.
\newblock {\em Proceedings of the 30th ACM International Conference on Multimedia}, 2022.

\bibitem{id-preserve}
S.-C. Lai, C.~He, and K.-M. Lam.
\newblock Low-resolution face recognition based on identity-preserved face hallucination.
\newblock {\em IEEE International Conference on Image Processing (ICIP)}, pages 1173--1177, 2019.

\bibitem{sc}
S.-C. Lai and K.-M. Lam.
\newblock Deep siamese network for low-resolution face recognition.
\newblock {\em Asia-Pacific Signal and Information Processing Association Annual Summit and Conference (APSIPA ASC)}, pages 1444--1449, 2021.

\bibitem{penb}
X.~Ling, Y.~Lu, W.~Xu, W.~Deng, Y.~Zhang, X.~Cui, H.~Shi, and D.~Wen.
\newblock Dive into the resolution augmentations and metrics in low resolution face recognition: {A} plain yet effective new baseline.
\newblock In {\em {AAAI} workshop}. {AAAI} Press, 2023.

\bibitem{cfsm}
F.~Liu, M.~Kim, A.~K. Jain, and X.~Liu.
\newblock Controllable and guided face synthesis for unconstrained face recognition.
\newblock In {\em Proc. of ECCV}, volume 13672, pages 701--719. Springer, 2022.

\bibitem{sphereface}
W.~Liu, Y.~Wen, Z.~Yu, M.~Li, B.~Raj, and L.~Song.
\newblock Sphereface: Deep hypersphere embedding for face recognition.
\newblock In {\em {IEEE} Conference on Computer Vision and Pattern Recognition, {CVPR}}, pages 6738--6746. {IEEE} Computer Society, 2017.

\bibitem{ideanet}
C.~Low and A.~B.~J. Teoh.
\newblock An implicit identity-extended data augmentation for low-resolution face representation learning.
\newblock {\em {IEEE} Trans. Inf. Forensics Secur.}, 17:3062--3076, 2022.

\bibitem{mindres}
C.-Y. Low, A.~Teoh, and J.~Park.
\newblock Mind-net: A deep mutual information distillation network for realistic low-resolution face recognition.
\newblock {\em IEEE Signal Processing Letters}, 28:354--358, 2021.

\bibitem{mobile}
Y.~Mart{\'i}nez-D{\'i}az, H.~M. Vazquez, L.~S. Luevano, L.~Chang, and M.~Gonz{\'a}lez-Mendoza.
\newblock Lightweight low-resolution face recognition for surveillance applications.
\newblock {\em International Conference on Pattern Recognition (ICPR)}, pages 5421--5428, 2021.

\bibitem{tc}
F.~V. Massoli, G.~Amato, and F.~Falchi.
\newblock Cross-resolution learning for face recognition.
\newblock {\em Image and Vision Computing}, 99:103927, 2020.

\bibitem{magface}
Q.~Meng, S.~Zhao, Z.~Huang, and F.~Zhou.
\newblock Magface: {A} universal representation for face recognition and quality assessment.
\newblock In {\em {IEEE} Conference on Computer Vision and Pattern Recognition, {CVPR}}, pages 14225--14234, 2021.

\bibitem{agedb}
S.~Moschoglou, A.~Papaioannou, C.~Sagonas, J.~Deng, I.~Kotsia, and S.~Zafeiriou.
\newblock Agedb: The first manually collected, in-the-wild age database.
\newblock In {\em {IEEE} Conference on Computer Vision and Pattern Recognition, {CVPR}}, 2017.

\bibitem{pytorch}
A.~Paszke, S.~Gross, S.~Chintala, G.~Chanan, E.~Yang, Z.~DeVito, Z.~Lin, A.~Desmaison, L.~Antiga, and A.~Lerer.
\newblock Automatic differentiation in pytorch.
\newblock In {\em NeurIPS Workshop}, 2017.

\bibitem{srdcr}
L.~Ren, W.~Deng, L.~Wang, C.~Mao, Y.~Jiang, H.~Jia, and J.~Li.
\newblock Low-resolution face recognition method combining super-resolution and improved dcr model.
\newblock {\em Proceedings of the 2020 9th International Conference on Computing and Pattern Recognition}, 2020.

\bibitem{cfp-fp}
S.~Sengupta, J.~Chen, C.~D. Castillo, V.~M. Patel, R.~Chellappa, and D.~W. Jacobs.
\newblock Frontal to profile face verification in the wild.
\newblock In {\em WACV}, 2016.

\bibitem{tcsvt-lq}
X.~Shan, Y.~Lu, Q.~Li, and Y.~Wen.
\newblock Model-based transfer learning and sparse coding for partial face recognition.
\newblock {\em IEEE Transactions on Circuits and Systems for Video Technology}, 31(11):4347--4356, 2021.

\bibitem{url}
Y.~Shi, X.~Yu, K.~Sohn, M.~Chandraker, and A.~K. Jain.
\newblock Towards universal representation learning for deep face recognition.
\newblock In {\em {IEEE/CVF} Conference on Computer Vision and Pattern Recognition, {CVPR}}, pages 6816--6825. {IEEE}, 2020.

\bibitem{coreface}
Y.~Song and F.~Wang.
\newblock Coreface: Sample-guided contrastive regularization for deep face recognition.
\newblock {\em CoRR}, abs/2304.11668, 2023.

\bibitem{catface}
N.~A. Talemi, H.~Kashiani, and N.~M. Nasrabadi.
\newblock Catface: Cross-attribute-guided transformer with self-attention distillation for low-quality face recognition.
\newblock {\em IEEE Transactions on Biometrics, Behavior, and Identity Science}, pages 1--1, 2024.

\bibitem{srda}
J.~C. Tan, K.~M. Lim, and C.~P. Lee.
\newblock Enhanced alexnet with super-resolution for low-resolution face recognition.
\newblock {\em International Conference on Information and Communication Technology (ICoICT)}, pages 302--306, 2021.

\bibitem{srgan}
M.~M.~I. Ullah, A.~Hamza, I.~A. Taj, and M.~Tahir.
\newblock Low resolution face recognition using enhanced srgan generated images.
\newblock {\em International Conference on Emerging Technologies (ICET)}, pages 1--6, 2021.

\bibitem{lrhr}
H.~Wang and S.~Wang.
\newblock Low-resolution face recognition enhanced by high-resolution facial images.
\newblock In {\em 2023 IEEE 17th International Conference on Automatic Face and Gesture Recognition (FG)}, pages 1--8, 2023.

\bibitem{cosface}
H.~Wang, Y.~Wang, Z.~Zhou, X.~Ji, D.~Gong, J.~Zhou, Z.~Li, and W.~Liu.
\newblock Cosface: Large margin cosine loss for deep face recognition.
\newblock In {\em {IEEE} Conference on Computer Vision and Pattern Recognition, {CVPR}}, pages 5265--5274. {IEEE} Computer Society, 2018.

\bibitem{wang_mis-classified_2020}
X.~Wang, S.~Zhang, S.~Wang, T.~Fu, H.~Shi, and T.~Mei.
\newblock Mis-classified vector guided softmax loss for face recognition.
\newblock In {\em The Thirty-Fourth {AAAI} Conference on Artificial Intelligence, {AAAI}}, pages 12241--12248. {AAAI} Press, 2020.

\bibitem{ijbb}
C.~Whitelam, E.~Taborsky, A.~Blanton, B.~Maze, J.~C. Adams, T.~Miller, N.~D. Kalka, A.~K. Jain, J.~A. Duncan, K.~Allen, J.~Cheney, and P.~Grother.
\newblock {IARPA} janus benchmark-b face dataset.
\newblock In {\em {IEEE} Conference on Computer Vision and Pattern Recognition, {CVPR}}, 2017.

\bibitem{improve-super}
L.~Xu and Z.~Gajic.
\newblock Improved network for face recognition based on feature super resolution method.
\newblock {\em International Journal of Automation and Computing}, 18:915 -- 925, 2021.

\bibitem{yang_orthogonality_2021}
S.-M. Yang, W.~Deng, M.~Wang, J.~Du, and J.~Hu.
\newblock Orthogonality loss: Learning discriminative representations for face recognition.
\newblock {\em {IEEE} Trans. Circuits Syst. Video Technol.}, 2021.

\bibitem{fan}
X.~Yin, Y.~Tai, Y.~Huang, and X.~Liu.
\newblock Fan: Feature adaptation network for surveillance face recognition and normalization.
\newblock In {\em Asian Conference on Computer Vision}, 2019.

\bibitem{tcn}
J.~Zha and H.~Chao.
\newblock Tcn: Transferable coupled network for cross-resolution face recognition*.
\newblock {\em IEEE International Conference on Acoustics, Speech and Signal Processing (ICASSP)}, pages 3302--3306, 2019.

\bibitem{super-id}
K.~Zhang, Z.~Zhang, C.-W. Cheng, W.~H. Hsu, Y.~Qiao, W.~Liu, and T.~Zhang.
\newblock Super-identity convolutional neural network for face hallucination.
\newblock In {\em Proc. of ECCV}, 2018.

\bibitem{texture}
M.~Zhang, R.~Liu, D.~Deguchi, and H.~Murase.
\newblock Texture-guided transfer learning for low-quality face recognition.
\newblock {\em IEEE Transactions on Image Processing}, 33:95--107, 2024.

\bibitem{cplfw}
T.~Zheng and W.~Deng.
\newblock Cross-pose lfw : A database for studying cross-pose face recognition in unconstrained environments.
\newblock In {\em Tech. Rep.}, 2018.

\bibitem{calfw}
T.~Zheng, W.~Deng, and J.~Hu.
\newblock Cross-age {LFW:} {A} database for studying cross-age face recognition in unconstrained environments.
\newblock {\em CoRR}, 2017.

\end{thebibliography}
}

\end{document}